\documentclass[runningheads]{llncs}
\usepackage{graphicx}
\usepackage{hyperref}
\usepackage[utf8]{inputenc}
\usepackage{amsmath}
\usepackage{amssymb}
\usepackage{fullpage} %I (CD) have commented this out to see how the paper looks like in the PPSN style. If you prefer fullpage for drafting, please just revert
\usepackage{xcolor}
\usepackage{todonotes}

\usepackage{ulem}
\usepackage{graphicx}
\usepackage{mathrsfs}
\usepackage{subcaption}
\usepackage{mathtools}
\usepackage{changepage}

\usepackage{algorithm}
\usepackage{algpseudocode}

\newcommand{\barX}{\ensuremath{\bar{\mathbf{X}}}}
\newcommand{\bx}{\mathbf{x}}

\newcommand{\af}{\mathscr{A}}         
 
\newcommand{\PEI}{\operatorname{PEI}} 
\newcommand{\var}{\operatorname{var}} 
\newcommand{\domain}{\mathrm{S}}

\DeclareMathOperator*{\argmax}{arg\,max}

\begin{document}

\title{High Dimensional Bayesian Optimization Assisted by Principal Component Analysis}
\titlerunning{Bayesian Optimization assisted by Principal Component Analysis}

\author{Elena Raponi\inst{1}%\orcidID{0000-0001-6841-7409}
\and Hao Wang\inst{2}%\orcidID{1111-2222-3333-4444} 
\and Mariusz Bujny\inst{3}%\orcidID{2222--3333-4444-5555}
\and Simonetta Boria\inst{1}%\orcidID{0000-0003-2073-4012} 
\and Carola Doerr\inst{2}%\orcidID{0000-0002-4981-3227}
}
\authorrunning{E. Raponi et al.}

\institute{University of Camerino, Via Madonna delle Carceri 9, 62032 Camerino, Italy \email{elena.raponi@unicam.it}
\and
Sorbonne Université, CNRS, LIP6, Paris, France
\and
Honda Research Institute Europe GmbH, %Carl-Legien-Straße 30, 
63073 Offenbach am Main, Germany}
\maketitle              % typeset the header of the contribution
\begin{abstract}

Bayesian Optimization (BO) is a surrogate-assisted global optimization technique that has been successfully applied in various fields, e.g., automated machine learning and design optimization. Built upon a so-called infill-criterion and Gaussian Process regression (GPR), the BO technique suffers from a substantial computational complexity and hampered convergence rate as the dimension of the search spaces increases. Scaling up BO for high-dimensional optimization problems remains a challenging task.\\
In this paper, we propose to tackle the scalability of BO by hybridizing it with a Principal Component Analysis (PCA), resulting in a novel PCA-assisted BO (PCA-BO) algorithm. Specifically, the PCA procedure learns a linear transformation from all the evaluated points during the run and selects dimensions in the transformed space according to the variability of evaluated points. We then construct the GPR model, and the infill-criterion in the space spanned by the selected dimensions.\\
We assess the performance of our PCA-BO in terms of the empirical convergence rate and CPU time on multi-modal problems from the COCO benchmark framework. The experimental results show that PCA-BO can effectively reduce the CPU time incurred on high-dimensional problems, and maintains the convergence rate on problems with an adequate global structure. PCA-BO therefore provides a satisfactory trade-off between the convergence rate and computational efficiency opening new ways to benefit from the strength of BO approaches in high dimensional numerical optimization.

\end{abstract}

\section{Introduction}
Over the last few years, Gaussian Process Regression (GPR)~\cite{kleijnen_kriging_2009} has been proven to be a very flexible and competitive tool in the modeling of functions that are expensive to evaluate or characterized by strong nonlinearities and measurement noises. 
A Gaussian Process (GP) is a stochastic process, i.e., a family of random variables, such that any finite collection of them have joint Gaussian distributions. GP is used in many application fields (engineering, geology, etc.) in order to evaluate datasets, predict unknown function values, and perform surrogate model-based optimization~\cite{forrester_engineering_2008}. Based on a certain number of evaluations of the objective function (training data), surrogate models allow for the construction of computationally cheap-to-evaluate approximations and for replacing the direct optimization of the real objective with the model. As such, many more evaluations can be performed on the approximate model.
GP uses a measure of the similarity between points -- the kernel function -- to predict the value for an unseen point from training data. GP is often chosen because it also provides a theoretical uncertainty quantification of the prediction error, among a variety of kernel-based methods, e.g., Radial Basis Functions~\cite{broomhead_radial_1988} and Support Vector Regression~\cite{vapnik_nature_1995,gunn_support_1998}.  Indeed, GP not only provides the estimate of the value of a function (the mean value), but also the variance at each domain point. This variance defines the uncertainty of the model while performing a prediction of the function value on an untested point of the search space. In particular, when GP is used in surrogate model-based optimization, at each iteration of the optimization algorithm, the estimate of the error is a crucial information in the update phase of the approximation model.\\
A GP can be used as a prior probability distribution over functions in
Bayesian Optimization (BO)~\cite{mockus1975bayesian,mockus2012bayesian}, also known as Efficient Global Optimization (EGO)
\cite{jones_efficient_1998}. BO is a sequential design strategy targeting black-box global optimization problems. It generates an initial set of training points in the search space where the objective function is evaluated. Based on this training data, a GPR model is built to approximate the real objective function. Afterwards, the optimization procedure starts by iteratively choosing new candidate points for which the objective function is evaluated. These points are obtained through the optimization of an acquisition function.  
However, it has to be noted that the optimization of the acquisition function and the model construction become quite complicated and time-consuming as the dimension of the problem increases, due to the well-known \textit{curse of dimensionality}. In particular, this is the case for most real-world optimization problems, where the number of design variables can range from a few (for simple parameter optimization problems) up to millions (for complex shape~\cite{hojjat2014vertex} or topology optimization~\cite{aage2015topology} problems). 
In spite of this difficulty, BO techniques have been successfully applied in many fields of engineering, including materials science~\cite{ueno2016combo,ju2017designing}, aerospace engineering~\cite{jeong2005efficient,forrester2006optimization,kanazaki2013mixed,lam2018advances}, turbomachinery design~\cite{huang2011optimal,arsenyev2015adaptive}, or even fluid and structural topology optimization~\cite{yoshimura_topology_2016,raponi_kriging-guided_2017,raponi_kriging-assisted_2019,raponi_hybrid_2019,liu_metamodel-based_2017}, thanks to a reduction of the problem dimensionality on the representation level.
Nevertheless, since the efficiency of these methods decreases strongly with the rising number of design variables, the development of BO approaches able to address high-dimensional problems is crucial for their future applications in industry~\cite{viana2014special,palar2019use}. 

Different dimensionality reduction techniques using other tools than PCA have been introduced in the literature for the optimization of expensive problems through GPs. 
Huang et al.~\cite{huang_scalable_2015} proposed a scalable GP for regression by training a neural network with a stacked denoising auto-encoder and performing BO afterwards on the top layer of the trained network. In~\cite{wang_bayesian_2016}, Wang et al. proposed to embed a lower-dimensional subspace into the original search space, through the so-called random embedding technique. Moreover, Blanchet-Scalliet et al.~\cite{blanchet-scalliet_four_2019} presented four algorithms to substitute classical kernels, which provide poor predictions of the response in high-dimensional problems. In~\cite{GaudrieRPEH2019}, a kernel-based approach is devised to perform the dimensionality reduction in a feature space for the shape optimization problem in Computer Aided
Design (CAD) systems. To the authors' knowledge, the only ones who used PCA to discover hidden features in GP regression are Vivarelli and Williams~\cite{vivarelli_discovering_1999}. However, their strategy mainly allowed for estimating which are the relevant directions in a prescribed feature space, rather than for pursuing the intrinsic optimization process. Finally, Kapsoulis et al.~\cite{kapsoulis_use_2016} embedded Kernel and Linear PCA in optimization strategies using Evolutionary Algorithms (EAs). Nevertheless, they identify the principal component directions and map the design space to a new feature space only when a prescribed iteration is reached, while in the first iterations PCA is not applied. In this paper, we propose to tackle this problem by using the well-known Principal Component Analysis (PCA) procedure to identify a proper subspace, on which we deploy the BO algorithm.
This treatment leads to a novel hybrid surrogate modeling method - the PCA-assisted Bayesian Optimization (PCA-BO), which starts with a Design of Experiments (DoE)~\cite{forrester_engineering_2008} and adaptively learns a linear map to reduce the dimensionality by a weighted PCA procedure. Here, the weighting scheme is meant for taking the objective values into account.
We assessed the empirical performance of PCA-BO by testing it on the well-known BBOB problem set~\cite{nikolaus_hansen_2019_2594848}. %which is well-recognized in the field of black-box optimization. 
Among these problems, we focus on the multi-modal ones, which are most representative of real-world optimization challenges. 

\section{Bayesian Optimization}
\label{sec:background}
In this paper, we consider the minimization of a real-valued objective function $f:\domain \subseteq \mathbb{R}^D \rightarrow \mathbb{R}$ with simple box constraints, i.e., $\domain = [\underline{\mathbf{x}}, \overline{\mathbf{x}}]$.
Bayesian Optimization~\cite{jones_efficient_1998,mockus2012bayesian} (BO) starts with sampling an initial DoE of size $n_0$: $\mathbf{X}=[\mathbf{x}_1,\mathbf{x}_2, \ldots, \mathbf{x}_{n_0}]^\top \subseteq \domain^{n_0}$. The DoE can be uniform random samples, or obtained through more sophisticated sampling methods~\cite{DBLP:books/daglib/0022270}. We choose the so-called Optimal Latin Hypercube Sampling (OLHS)~\cite{fang_design_2005} in this paper. The corresponding objective function values are denoted as $\mathbf{y}=(f(\mathbf{x}_1),f(\mathbf{x}_2),\ldots,f(\mathbf{x}_{n_0}))^\top$. Conventionally, a centered Gaussian process prior is assumed on the objective function: $f \sim gp(0, k(\cdot, \cdot))$, where $k\colon \domain \times \domain \rightarrow \mathbb{R}$ is a positive definite function (a.k.a.~\textit{kernel}) that computes the autocovariance of the process. 
Often, a Gaussian likelihood is taken, leading to a conjugate posterior process~\cite{rasmussen2006gaussian}, i.e., $f \;|\; \mathbf{y} \sim gp(\hat{f}(\cdot), k'(\cdot, \cdot))$, where $\hat{f}$ and $k'$ are the posterior mean and covariance function, respectively.
On an unknown point $\mathbf{x}$, $\hat{f}(\mathbf{x})$ yields the maximum a posteriori (MAP) estimate of $f(\mathbf{x})$ whereas $\hat{s}^2(\mathbf{x}) \coloneqq k'(\mathbf{x}, \mathbf{x})$ quantifies the uncertainty of this estimation. We shall refer to the posterior process as the \emph{Gaussian process regression} (GPR) model. Based on the posterior process, we usually identify promising points via the so-called \textit{infill-criterion} which %that typically 
balances $\hat{f}$ with $\hat{s}^2$ (exploitation vs. exploration). A variety of infill-criteria has been proposed in the literature, e.g., Probability of Improvement~\cite{forrester_engineering_2008,mockus2012bayesian}, Expected Improvement~\cite{forrester_engineering_2008}, and the Moment-Generating Function of Improvement~\cite{DBLP:conf/smc/0025SEB17}. In this work, we adopt the Penalized Expected Improvement (PEI) criterion~\cite{raponi_kriging-assisted_2019} to handle the box constraints,
%\hw{@Elena: Could you add the penalty term?}
\begin{equation}\text{PEI}(\bx) =
\begin{cases}
E\{\max\{0, \min \mathbf{y} - f(\mathbf{x})\} \;|\; \mathbf{y}\} &\mbox{if $\bx$ is feasible}, \\
-P(\bx)  &\mbox{if $\bx$ is infeasible},
\end{cases}
\label{eq:PEI}
\end{equation}
which essentially calculates the expected improvement when $\mathbf{x}$ is feasible, and returns a negative penalty value otherwise. The penalty function $P(\mathbf{x})$ is proportional to the degree of infeasibility of the point $\mathbf{x}$. This will be further described in Sec.~\ref{sec:method}.
%\hw{Elena, $P(\mathbf{x})$ is the penalty function, correct? We need to define it here.} 
BO then proceeds to select the next point by maximizing PEI, namely, $\mathbf{x}^* = \argmax_{\mathbf{x}\in\domain} \PEI(\mathbf{x})$. After evaluating $\mathbf{x}^*$, BO augments\footnote{With an abuse of terminology, the operation $\mathbf{X} \cup\{\mathbf{x}^*\}$ is understood as appending $\mathbf{x}^*$ at the bottom row of $\mathbf{X}$ throughout this paper. $\mathbf{y}\cup\{f(\mathbf{x}^*)\}$ is defined similarly.} the data set, $\mathbf{X} \leftarrow \mathbf{X} \cup\{\mathbf{x}^*\}, \mathbf{y} \leftarrow \mathbf{y}\cup\{f(\mathbf{x}^*)\}$, on which the GPR model is retrained.

\section{PCA-assisted Bayesian Optimization}
\label{sec:method}
In this section, we introduce the coupling of the principal component analysis procedure and Bayesian optimization method, called \textbf{PCA-BO}. Loosely speaking, this approach learns a linear transformation (from the initial design points of BO) that would, by design, identify directions (a.k.a.~principal components) in $\mathbb{R}^D$ along which the objective value changes rapidly. Also, this approach adapts this transformation through the optimization process. Intuitively, we could further drop the components to which objective function values are less sensitive, resulting in a lower-dimensional search space $\mathbb{R}^r$ ($r<D$). Particularly for BO, the surrogate model shall be trained in this lower-dimensional search space, thus benefiting from the dimensionality reduction. Also, the infill-criterion optimization, which is typically an expensive sub-procedure in BO, becomes less costly because it now operates in the lower-dimensional space as well. 

\begin{figure}[!htb]
\centering
\includegraphics[width=1\textwidth,clip]{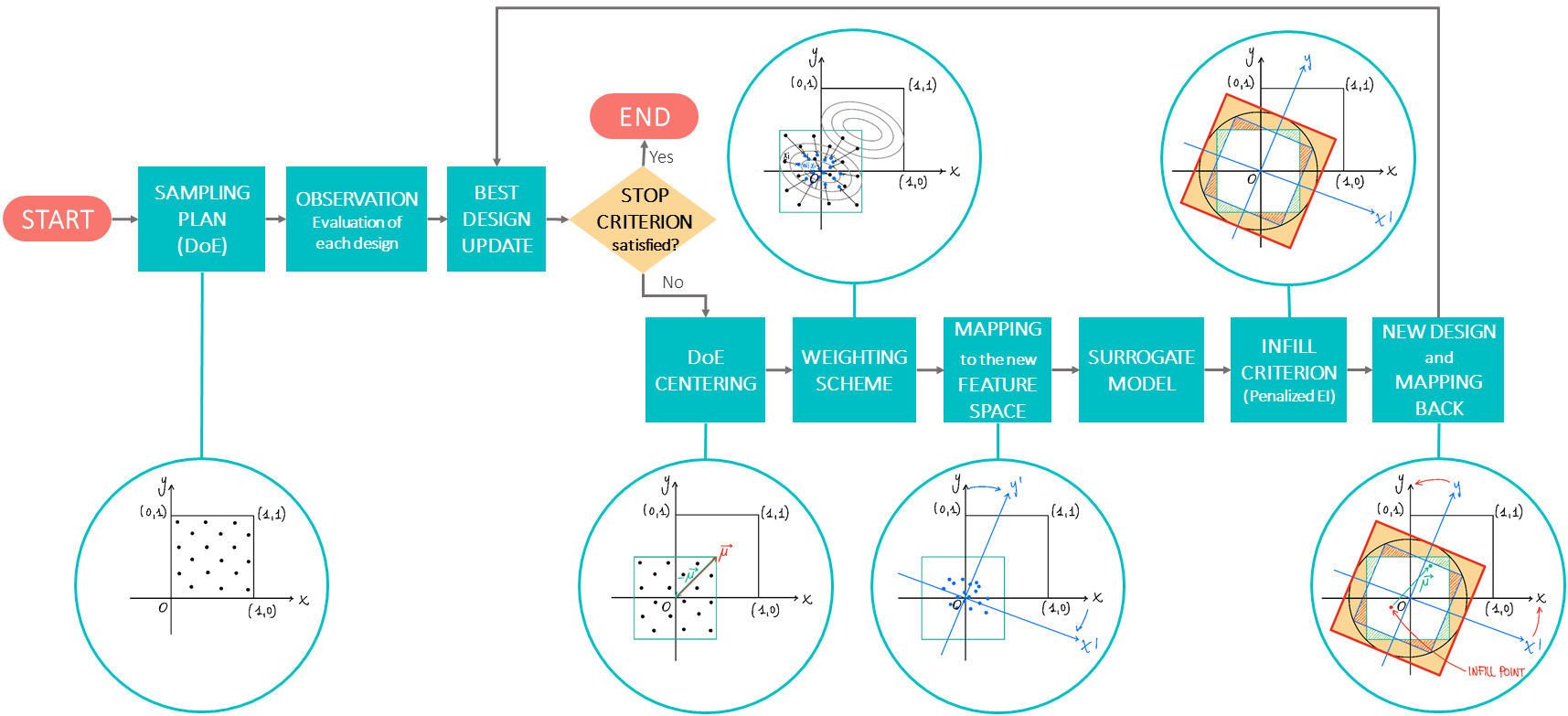}
\caption{Flowchart of the PCA-BO optimization algorithm in which PCA-related operations are depicted in circles.}
\vspace{-6mm}
\label{fig:flowchart}
\end{figure}

\paragraph{The PCA-BO algorithm.}
% \label{subsec:PCABO_strategy}
We first present a high-level overview of the proposed PCA-BO algorithm. A graphical representation is provided in Fig.~\ref{fig:flowchart}, whereas a pseudo-code description is presented in Alg.~\ref{alg:pcabo}.
PCA-BO works as follows:

\begin{enumerate}
    \item We perform an initial DoE in the original search space, generating as an outcome a set of evenly distributed points. 
    \item We design a weighting scheme to embed the information from objective function into the DoE points, where smaller weights are assigned to the points with worse function values.
    \item We apply a PCA procedure to obtain a linear map from the original search space to a lower-dimensional space, using the weighted DoE points.
    \item In the lower-dimensional space, we train a GPR model and maximize an infill-criterion to find a promising candidate point.
    \item We map the candidate point back to the original search space, and evaluate it with the objective function.
    \item We augment the data set with the candidate point and its objective value, and then proceed to step 2 for the next iteration.
\end{enumerate}
\vspace{-5mm}

\begin{algorithm}[!ht]
	\caption{PCA-assisted Bayesian Optimization}\label{alg:pca-bo}
	\begin{algorithmic}[1]
		\Procedure{pca-bo}{$f,\af,\domain$} 
		\Comment{$f$: objective function, $\af$: infill-criterion, $\domain$: search space, $\domain'$: lower-dimensional search space}
		\State{Create $\mathbf{X} = [\mathbf{x}_1,\mathbf{x}_2,\ldots,\mathbf{x}_{n_0}]^\top \subset \domain^{n_0}$ with optimal Latin hypercube sampling}
		\State{$\mathbf{y} \leftarrow (f(\mathbf{x}_1),f(\mathbf{x}_2),\ldots,f(\mathbf{x}_{n_0}))^\top, \; n \leftarrow n_0$}
		
		\While{the stop criteria are not fulfilled}
		    \State{Centering: $\barX \leftarrow  \mathbf{X} - \mathbf{1}_n \boldsymbol{\mu}$}
		    \State{Rescaling: $\mathbf{X'} \leftarrow \barX\mathbf{W} $}
		    \State{$\boldsymbol{\mu'}, \mathbf{P}_r \leftarrow \textsc{pca}(\mathbf{X'})$}
		  %  \State{Apply \textsc{pca} to $\mathrm{X}'$ to learn the linear transformation of parameters $\boldsymbol{\mu'}$ and $\mathrm{P_r}$}
		    \State{Mapping to $\mathbb{R}^r$: $\mathbf{Z}_r \leftarrow \mathbf{P}_r(\barX - \mathbf{1}_n\boldsymbol{\mu}')^\top$}
		    \State{GPR training: $\hat{f}, \hat{s}^2 \leftarrow \textsc{gpr}(\mathbf{Z}_r,\mathbf{y})$}
	 		\State{$\mathbf{z}' \leftarrow \argmax_{\mathbf{z}\in \domain'}\af(\mathbf{z};\hat{f}, \hat{s}^2)$}
	 		\State{Mapping to $\mathbb{R}^D$: $\mathbf{x}' \leftarrow \mathbf{P}_r^{\top}\mathbf{z} + \boldsymbol{\mu'} + \boldsymbol{\mu}$}
	 		\State{$y' \leftarrow f(\mathbf{x}')$}
	 		\State{$\mathbf{X} \leftarrow \mathbf{X} \cup \{\mathbf{x}'\}, \;\mathbf{y} \leftarrow (\mathbf{y}^\top, y')^\top, \; n\leftarrow n + 1$}
	 	\EndWhile
	 	\EndProcedure
	\end{algorithmic}
	\label{alg:pcabo}
\end{algorithm}

\paragraph{Rescaling data points for PCA.}
Because the goal of applying PCA, as stated above, is to spot directions in $\mathbb{R}^D$ to which the function value is sensitive, it is then necessary to take into account the information on the objective function. We implement this consideration by rescaling the data points according to a weighting scheme, which depends on the corresponding function values. The weighting scheme aims to adjust the ``importance'' of data points when applying the PCA procedure, such that a point associated with a better (here lower) function value is assigned with a larger weight.
We propose to use a \emph{rank-based weighting scheme}, in which the weight assigned to a point is solely determined by the rank thereof. Let $\{r_i\}_{i=1}^n$ be the ranking of the points in $\mathbf{X}$ according to their function values $\mathbf{y}$. We calculate the rank-based (pre-)weights as follows:
%-----------------------------
\begin{equation} \label{eq:rank-weighting}
\tilde{w}_i = \ln{n} - \ln{r_i},
\end{equation}
%-----------------------------
where $n$ stands for the current number of data points. Afterwards, we normalize the pre-weights: $w_i = \tilde{w}_i / \sum \tilde{w}_i$. Notably, this weighting scheme will gradually decrease the weights of worse points when a better point is found and added to the data, hence leading to a self-adjusting discount factor on the data set.\footnote{Alternatively, the weight can also be computed directly from the function value, e.g., through a parameterized hyperbolic function. However, we do not  prefer this approach since it introduces extra parameters that require tuning, and does not possess the discount effect of the rank-based scheme since the weights remain static throughout the optimization.} It is worth pointing out that this weighting scheme resembles the weight calculation in the well-known Covariance Matrix Adaptation Evolution Strategy (CMA-ES)~\cite{Hansen2006}, although they are motivated differently. 
For ease of discussion, we will use a weight matrix $\mathbf{W} = \operatorname{diag}(w_1, w_2, \ldots, w_n)$ henceforth. We rescale the original data $\mathbf{X}$ by removing its sample mean, i.e., $\bar{\mathbf{X}} = \mathbf{X} - \mathbf{1}_n\boldsymbol{\mu}$, 
where
$
\boldsymbol{\mu} = \left(\frac{1}{n} \sum_{i=1}^n X_{i1}, \ldots, \frac{1}{n} \sum_{i=1}^n X_{id}\right) \text{ and } \mathbf{1}_n = (1,\ldots,1)^\top
$ ($n$ 1's),
and then adjust each point with the corresponding weight, namely, $\mathbf{X}' = \bar{\mathbf{X}}\mathbf{W}$.

\paragraph{Dimensionality reduction.} The PCA procedure starts with computing the sample mean of data matrix $\mathbf{X}'$, i.e., $\boldsymbol{\mu}' = \left(\frac{1}{n} \sum_{i=1}^n X_{i1}', \ldots, \frac{1}{n} \sum_{i=1}^n X_{id}'\right)$, after which it calculates the unbiased sample covariance matrix $\mathbf{C} = \frac{1}{n-1}{\mathbf{X}''}^\top{\mathbf{X}''}, \mathbf{X}'' = \mathbf{X}' - \mathbf{1}_n\boldsymbol{\mu}'.$
The principle components (PCs) are computed via the eigendecomposition $\mathbf{C} = \mathbf{P}\mathbf{D}\mathbf{P}^{-1}$, where column vectors (a.k.a.~eigenvectors) of $\mathbf{P}$ identify the PCs, and the diagonal matrix $\mathbf{D}=\operatorname{diag}(\sigma^2_1,\ldots,\sigma^2_d)$ contains the variance of $\mathbf{X}''$ along each principal component, i.e., $\sigma^2_i = \var\{\mathbf{X}''\mathbf{p}_i\}$ ($\mathbf{p}_i$ is the $i^{th}$ column of $\mathbf{P}$). For clarity of discussion, we consider the orthonormal matrix $\mathbf{P}$ that defines a rotation operation, as a change of the standard basis in $\mathbb{R}^D$ such that PCs are taken as the basis after the transformation. After changing the basis, we proceed to reduce the dimensionality of $\mathbb{R}^D$ by sorting the PCs according to the decreasing order of their variances and keeping the first $r$ components. Here~$r$ is chosen as the smallest integer such that the top-$r$ variances sum up to at least $\alpha$~percent of the total variability of the data (we use $\alpha = 95\%$ in our experiments). By denoting the ordering of variances and PCs as $\sigma^2_{1:d},\sigma^2_{2:d}\ldots,\sigma^2_{d:d}$ and $\mathbf{p}_{1:d}, \mathbf{p}_{2:d},\ldots, \mathbf{p}_{d:d}$, respectively, we formulate the selection of PCs as follows:
%-----------------
$$\mathbf{P}_r \coloneqq [\mathbf{p}_{1:d}, \mathbf{p}_{2:d},\ldots, \mathbf{p}_{r:d}]^\top \in\mathbb{R}^{r\times d},\; r=\inf\left\{k\in[1..d]\colon\sum_{i=1}^k \sigma^2_{i:d} \geq \alpha \sum_{i=1}^{d} \sigma^2_i\right\}.$$
%-----------------
Now, we could transform the data set $\mathbf{X}$ into a lower-dimensional space, i.e., $\mathbf{Z}_r =  \mathbf{P}_r(\bar{\mathbf{X}}- \mathbf{1}_n\boldsymbol{\mu}')^\top$. Note that, 1) in this transformation, the centered data set $\bar{\mathbf{X}}$ is not scaled by the weights because we only intend to incorporate the information on the objective values when determining $\mathbf{P}_r$. Using the scaled matrix $\bar{\mathbf{X}}\mathbf{W}$ will make it cumbersome to define the inverse mapping (see below). 2) It is crucial to substract $\boldsymbol{\mu}'$ from $\bar{\mathbf{X}}$ since it estimates the center of ellipsoidal contours when $f$ resembles a quadratic function globally. In this way, the principal axis of contours will be parallel to the PCs after applying $\mathbf{P}_r$. 3) Here, $\mathbf{P}_r\colon \mathbb{R}^D \rightarrow \mathbb{R}^r$ is not injective ($\operatorname{dim}(\operatorname{ker}\mathbf{P}_r) = D - r$). Therefore when transforming the centered data set to the lower-dimensional space, it seems that we might loose some data points in $\barX$. However, we argue that such an event $\{\mathbf{x} - \mathbf{x}'\in \operatorname{ker}\mathbf{P}_r\colon \mathbf{x},\mathbf{x}' \in \mathbf{X}\}$ is a null set with respect to any non-singular measure on $\mathbb{R}^D$, and the probability distribution of $\mathbf{X}$, no matter which forms it takes, should not be singular (otherwise BO will only search in a subspace of $\mathbb{R}^D$). Thus, it is safe to ignore the possibility of losing data points.
% \hw{@Elena: I think it is measure zero.. let discuss this point tomorrow and find an argument for it.}
Now we are ready to train a GPR model on $(\mathbf{Z}_r,\mathbf{y})$, and propose candidate points by optimizing an infill-criterion in $\mathbb{R}^r$. Importantly, to evaluate a candidate point $\mathbf{z}\in\mathbb{R}^r$, we map it back to $\mathbb{R}^D$ on which the objective function takes its domain, using the following linear map $L$:
%-----------------
%\forall \mathbf{z} \in \mathbb{R}^r, \quad
$$ L\colon\mathbb{R}^r \rightarrow \mathbb{R}^D, \quad\mathbf{z} \mapsto \mathbf{P}_r^{\top}\mathbf{z} + \boldsymbol{\mu'} + \boldsymbol{\mu}.$$
%-----------------
Note that, 1) the linear map $L$ is an injection; 2) it is necessary to include $\boldsymbol{\mu}$ and $\boldsymbol{\mu}'$ in this transformation since we subtract them from the design matrix $\mathbf{X}$ when mapping it to the lower-dimensional space; 3) After evaluating point $\mathbf{z}$, i.e., $y=f(L(\mathbf{z}))$, we augment the data set in $\mathbb{R}^D$: $\mathbf{X}\cup \{L(\mathbf{z})\}$ and $\mathbf{y}\cup \{y\}$, from which the linear map $\mathbf{P}_r$ is re-computed using the procedure described above; 4) After re-computing $\mathbf{P}_r$, we have to re-calculate $\mathbf{Z}_r$ by mapping the augmented $\mathbf{X}$ into $\mathbb{R}^r$: $\mathbf{Z}_r = \mathbf{P}_r(\bar{\mathbf{X}}- \mathbf{1}_n\boldsymbol{\mu}')^\top$, and retrain the GPR model on $(\mathbf{Z}_r, \mathbf{y})$.

\paragraph{Penalized infill-criterion.}
Once the regression model is fitted, we use it to construct and maximize the acquisition function that determines the next infill point. 
As previously stated, in this work we use a penalized version of the well-known EI, referred to as PEI. By definition, we search for the maximum of PEI through DE over a prescribed cubic domain $\mathrm{C}$, which is orientated according to the axes of the transformed coordinate system and must contain the projection  of the PCA-transform of the old domain $\mathrm{S}$ onto the lower-dimensional feature space.
We hence define and penalize a bounding cube $\mathrm{C}$ in order to prevent 1) that the optimum is found in a region of $\mathrm{C}$ that falls outside $\mathrm{S}$ when mapped back, and 2) that we neglect some good regions for new query points, i.e., the regions that belong $\mathrm{S}$ but not to $\mathrm{C}$.
To achieve this, if $\rho$ is the radius of the sphere tangent to $\mathrm{S}$, we define 
$
\mathrm{C} = [C_1 - \rho, C_1 + \rho] \times \dots \times [C_r - \rho, C_r + \rho],
$
where $C_i$, $i = 1, \dots, r$ is the $i^\text{th}$ component of the center of the new feature space. 
Since, for simplicity reasons, $\mathrm{C}$ is wider than necessary, we introduce a penalty to be assigned to the points that would not fall into $\mathrm{S}$ when mapped back through the inverse PCA linear transformation -- the \textit{infeasible points}. 
In order to avoid a stagnation of the DE search, the penalty $P$ assigned to infeasible points is defined as the additive inverse of the distance of their images through the PCA inverse transformation from the boundary of $\mathrm{S}$.
As a result, the infeasible points are automatically discarded in the maximization of PEI (Eq.~\eqref{eq:PEI}).

\section{Experiments}
\label{sec:exp}

\paragraph{Experimental Setup.}
We assess the performance of PCA-BO on ten multi-modal functions taken from the the BBOB problem set~\cite{nikolaus_hansen_2019_2594848}, and compare the experimental result to a standard BO.
This choice of test functions aims to verify the applicability of PCA-BO to more difficult problems, which potentially resemble the feature of real-world applications.
We test PCA-BO on three different dimensions $D\in\{10, 20, 40\}$ with the following budget of $10D + 50$ function evaluations, of which the initial DoE takes up to 20\%. 
To allow for a statistically meaningful comparison, we performed $30$ independent runs for each combination of algorithm, function, and dimension, on an Intel(R) Xeon(R) CPU E5620 @ 2.40GHz machine in single-threaded mode. Also, we recorded the best-so-far function value and measured CPU time taken by those algorithms, which are considered as performance metrics in this paper.
We select a squared-exponential kernel function for the GPR model that underpins both PCA-BO and BO. A Differential Evolution (DE)~\cite{storn_differential_1997} algorithm\footnote{We take the \texttt{Scipy} implementation of DE (\url{https://docs.scipy.org/doc/scipy/reference/generated/scipy.optimize.differential_evolution.html}) with a population size of $20r$ and the ``best1bin'' strategy, which uses the binary crossover and calculates the differential vector based on the current best point. Here, we set the evaluation budget to $20020r^2$ to optimize the infill-criterion.} is adopted to maximize the PEI criterion in every iteration.
We implement both algorithms in \texttt{Python}, where the \texttt{PyDoE}, \texttt{PyOpt}, and \texttt{Scikit-learn} open-source packages are used to handle the DoE, modeling, and PCA-transformation phases, respectively. 

\paragraph{Results.}
% \label{subsec:result}
In this section, we illustrate the results we obtained in the optimization of multi-modal benchmark functions from the BBOB suite by using PCA-BO and the state-of-the-art BO. In particular, each algorithm is tested on functions \textbf{F15-F24}, which can be categorized in two main groups: 
1) \emph{multi-modal functions with adequate global structure}: \textbf{F15-19}, and 2) \emph{multi-modal functions with weak global structure}: \textbf{F20-24}. By keeping $95\%$ of the variability of data points, our PCA-assisted approach, on average, reduces about 40\% dimensions from the original search space across all problems and dimensions. Also, we compare the two algorithms in terms of the empirical convergence and CPU time.
%-----------------------------------------------
\begin{figure}[!htbp]
\centering
\includegraphics[width=1\textwidth, trim=0mm 35mm 0mm 00mm, clip]{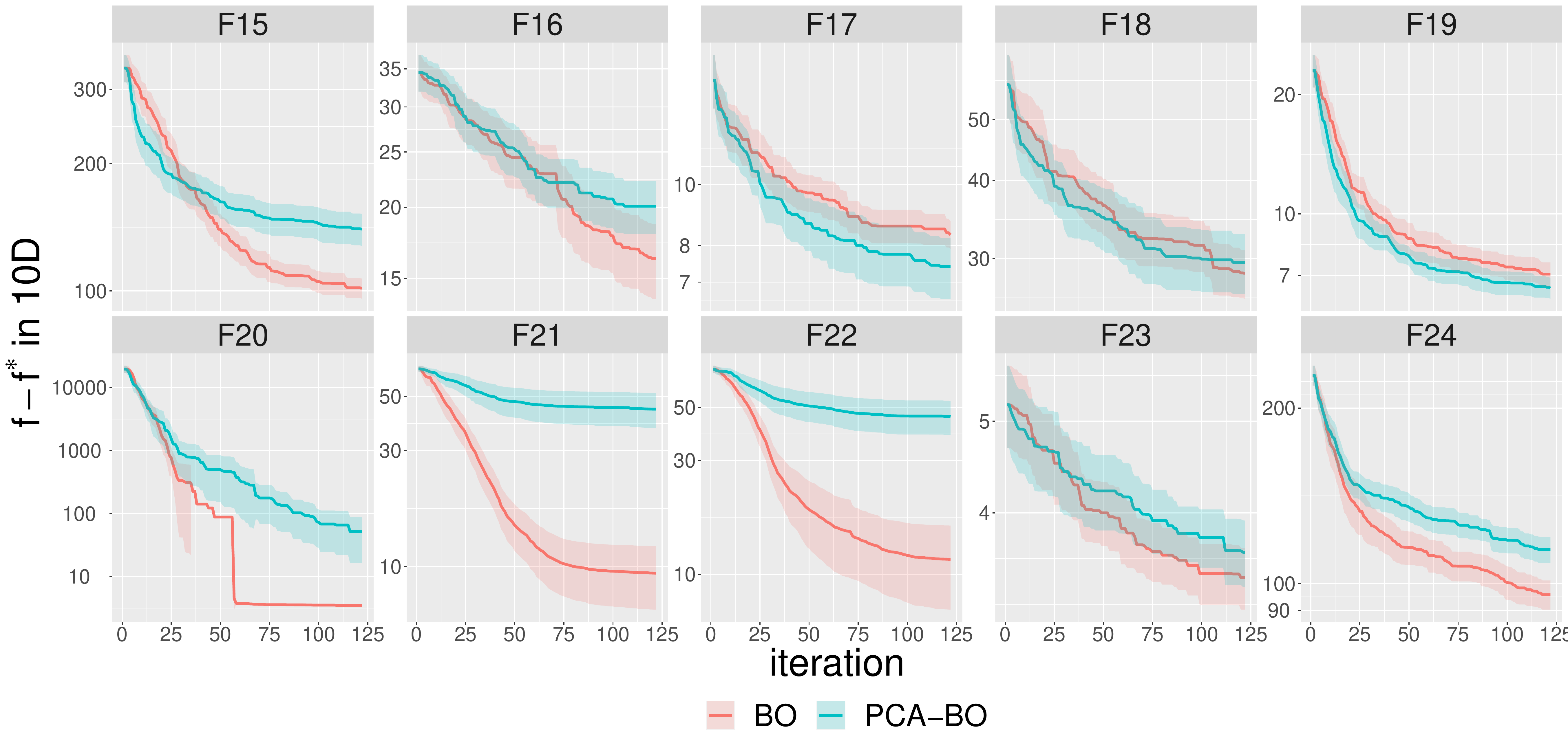} \\
\includegraphics[width=1\textwidth, trim=0mm 35mm 0mm 00mm, clip]{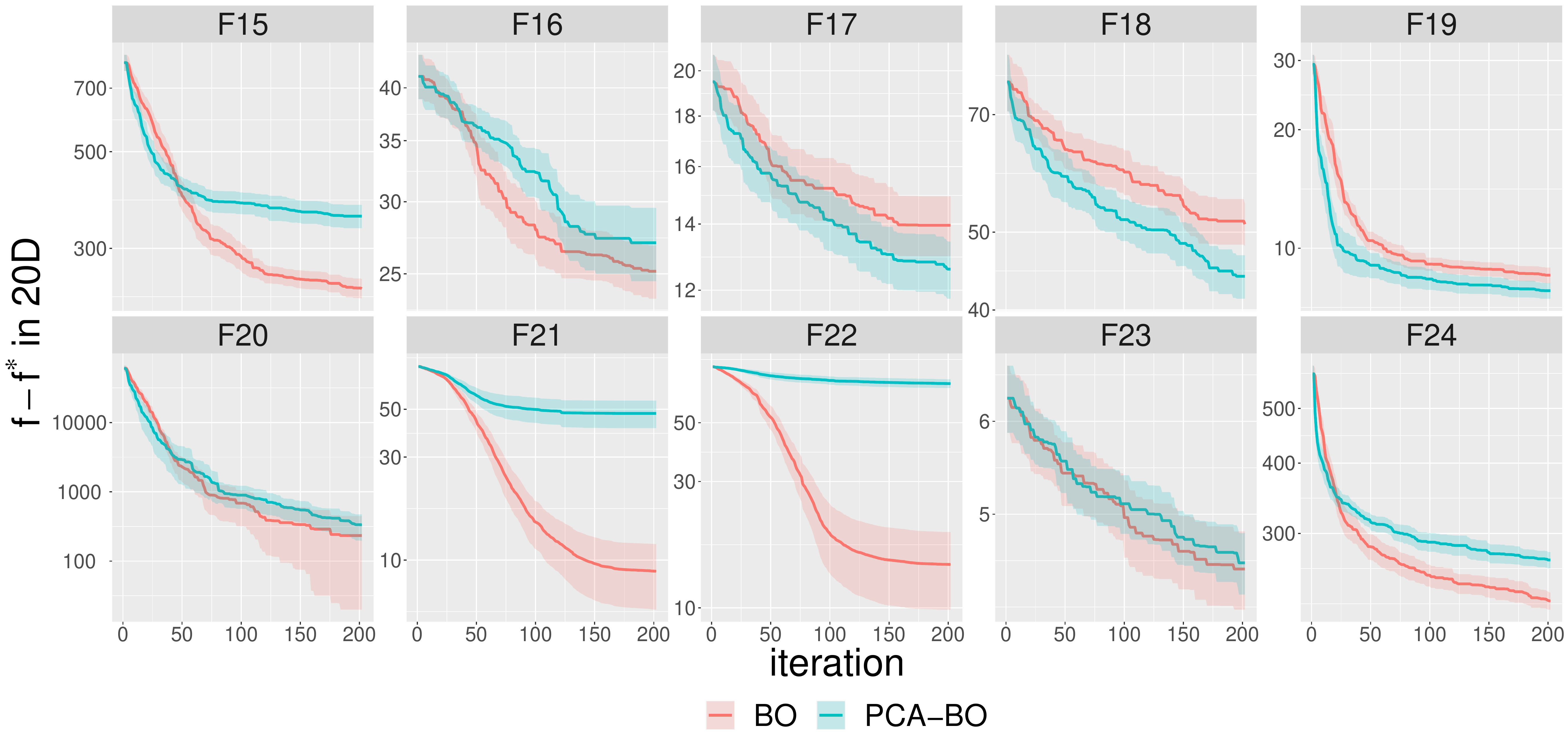} \\
\includegraphics[width=1\textwidth]{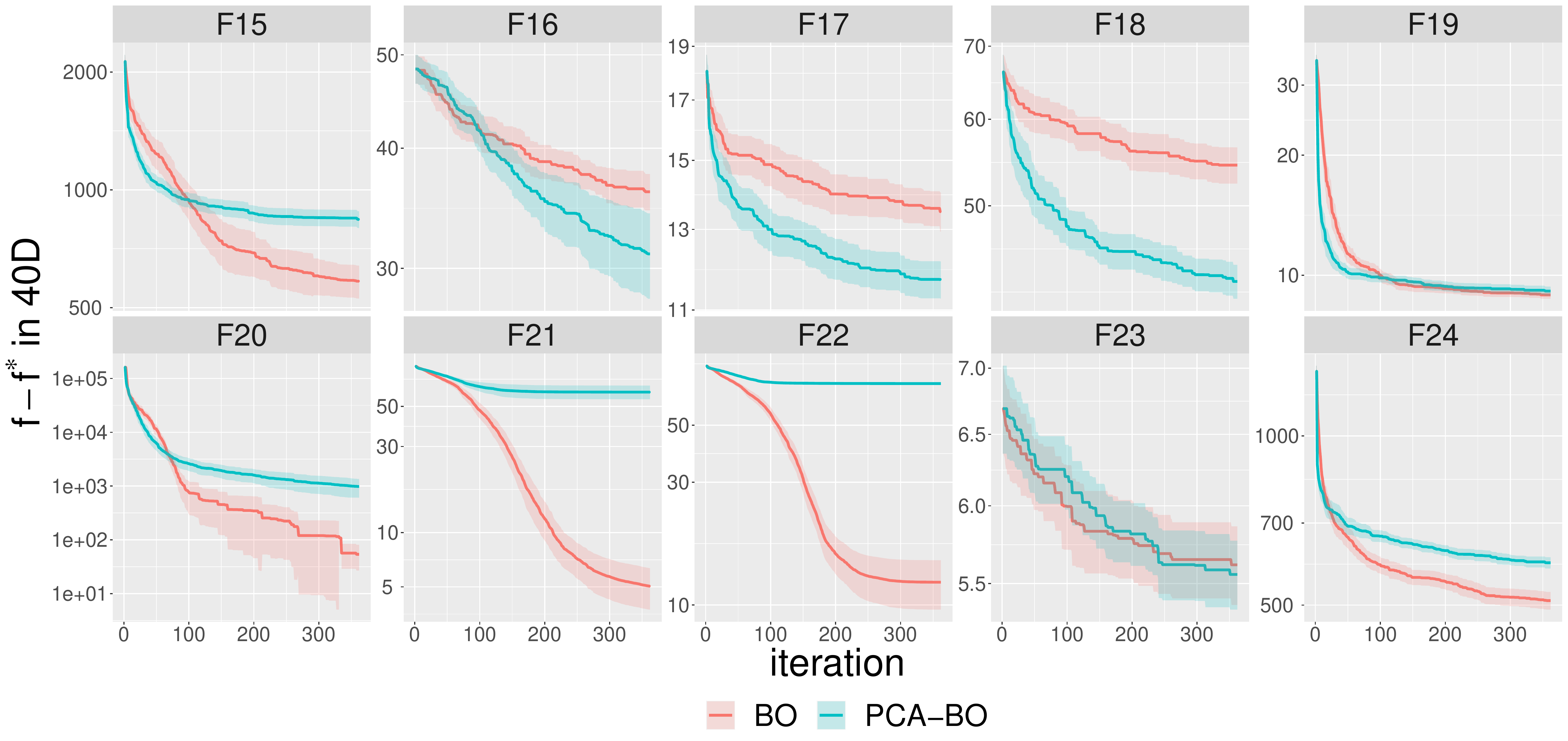} 
\caption{Mean target precision to the optimal value against the iteration for BO (red) and PCA-BO (blue), on multi-modal functions F15-24 from the BBOB problem set, in three dimensionalities, $10$D (top), $20$D (middle), and $40$D (bottom). $30$ independent runs are conducted and the shade area indicates the $95\%$ confidence interval of the mean target precision. The iteration counts the number of function evaluations after evaluating the initial design points.}
\label{fig:convergence_plots}
\end{figure}
%-------------------------------------------
\begin{figure}[!ht]
\centering%
% \begin{adjustwidth}{-1cm}{-.08cm}
\includegraphics[width=1\textwidth, trim=15mm 5mm 0mm 0mm,clip]{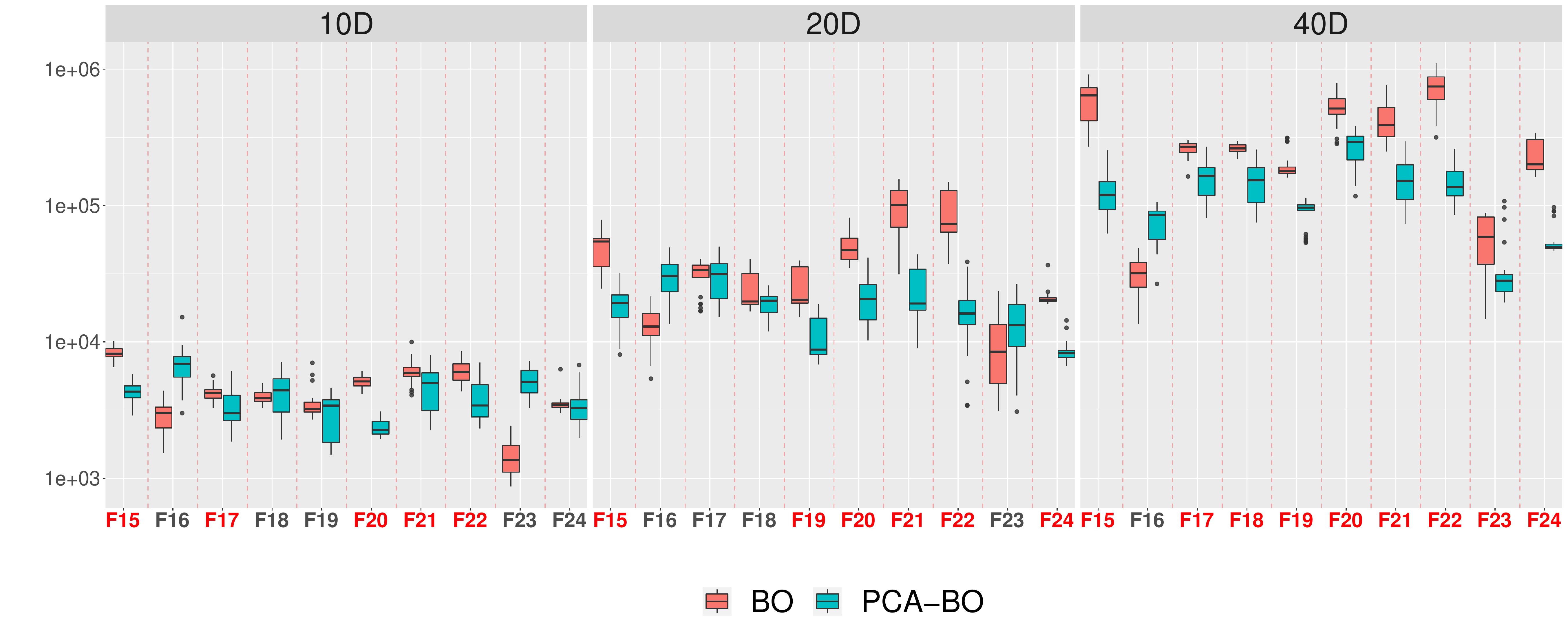}
% \end{adjustwidth}
\caption{CPU time (in seconds) taken by PCA-BO and BO to complete the evaluation budget for all functions and three dimensions. For each pair of problems and dimensions, we also applied the well-known Mann–Whitney U test on the CPU time with the Holm–Bonferroni correction of $p$-values and a significance level of $0.01$. The red-colored test problems indicate that the CPU time of PCA-BO is significantly smaller than BO based on the test.
\label{fig:boxplots}}
\vspace{-3mm}
\end{figure}
In Fig.~\ref{fig:convergence_plots}, we report the mean target precision achieved on each test function, which is the difference between the best-so-far function value and the optimal function value averaged over 30 runs with the same initial DoE points. Also, we plot the mean target precision against the iteration number that counts the number of function evaluations after the DoE phase and illustrate the $95\%$ confidence interval of the mean target precision.\footnote{On the 10-dimensional \textbf{F20} problem, we observed that the standard deviation of BO over 30 runs gradually shrinks to zero after $50$ iterations, making the confidence interval disappear in the corresponding subplot.} 
On each dimension, the plots clearly illustrate that PCA-BO is either superior or at least comparable to the standard BO when the test function exhibits an adequate global structure (\textbf{F15-19}). For each function in this group, PCA-BO reveals a steep convergence at the beginning of the optimization procedure. Notably, such an early and steep convergence is extremely valuable when we apply it in real-world scenarios, e.g., industrial optimizations, where we could only afford a very limited evaluation budget for expensive optimization problems. 
In contrast, on multi-modal functions with weak global structure (\textbf{F20-24}), PCA-BO has difficulties in maintaining the convergence speed, particularly on functions F21 and F22 whose global contour lines are composed of multiple local ellipsoids that have different orientations (see contour plots in~\cite{nikolaus_hansen_2019_2594848}). We speculate that the poor performance of PCA-BO on those two functions is because we apply the PCA procedure on a global scale, which could be potentially confused by the landscape of F21 and F22.
However, we argue that this issue could be alleviated effectively with a simple extension of applying the PCA procedure locally, which can be realized by either searching for the optimum with the trust region approach~\cite{yuan_review_nodate}, or through clustering techniques~\cite{van_stein_cluster-based_2017}. We shall investigate this issue in future work.
Moreover, we performed the well-known Wilcoxon rank-sum test on the target precision values reached by each algorithm upon termination in each dimension.
Under the null hypothesis of this test, the target precision values has an equal distribution in BO and PCA-BO~\cite{MATLAB:2018}. Based on the data, we reject the null hypothesis at the $5\%$ significance level on all functions and dimensions except F23, which belongs to the category of multi-modal functions with weak global structure. 
Therefore, we argue that PCA-BO converges at least as fast as the standard BO algorithm except on problems \textbf{F20}-\textbf{F24}, where the problem landscape exhibits high local irregularities and lacks a global structure. This scenario will be addressed by the authors in future works. More importantly, the PCA-BO algorithm achieves this convergence speed with a significantly less computational power, which is supported by measured CPU times in Fig.~\ref{fig:boxplots}. Here, we observe that the CPU time of PCA-BO, on average, is about $1.2209$ times that of BO on $10$D. Although this ratio does not imply any benefits from PCA-BO on $10D$, it, however, decreases to $0.7519$ for $20$D, and further to $0.5910$ for $40$D. Therefore, PCA-BO can effectively reduce the CPU time incurred on higher-dimensional problems, leading to a $25\%$ and a $41\%$ reduction in the 20- and 40-variables tests, respectively.  

\section{Conclusions and future research}
\label{sec:Conclusions}
In this paper, we presented the novel PCA-assisted Bayesian Optimization (PCA-BO) method, which aims to efficiently address high-dimensional black-box optimization problems by means of surrogate modeling techniques. In fact, despite BO has been successfully applied in various fields, it demonstrated to be restricted to problems of moderate dimension, typically up to about 15. Therefore, since many real-word applications require the solution of very high-dimensional optimization problems, it becomes necessary to scale BO to more complex parameter spaces, e.g., through dimensionality reduction techniques. 
Essentially, PCA-BO allows for achieving this by performing iteratively an orthogonal linear transformation of the original data-set, in order to map it to a new feature space where the problem becomes \textit{more separable} and features are selected in order to inherit the maximum possible variance from the original sample set. Once the transformation is applied, the model construction and the optimization of the acquisition function -- i.e., the costly steps characterizing the BO sequential strategy -- are performed in the lower-dimensional space so that we can find a promising point by saving computational resources. Afterwards, the selected point is mapped back to the original feature space, it is evaluated, added to the sample set to increase the accuracy of the model, and compared to the previous points to eventually update the optimum.\\ 
We empirically evaluated PCA-BO by testing it on the well-known COCO/BBOB benchmark suite. In particular, we used the presented method to optimize two sets of multi-modal benchmarks, characterized by either an adequate or weak global structure. We also ran the standard BO for comparison purposes. 
It is clear from the results that PCA-BO achieves state-of-the-art BO performance for optimizing multi-modal benchmarks with an adequate global structure, while it finds some difficulties when dealing with multi-modal benchmarks with a weak global structure. We ascribe such inefficiency to the global nature of the PCA-BO algorithm. In fact, when learning the the orthogonal linear transformation from all the evaluated points during the run, the proposed algorithm cannot detect a clear and well-definite trend of the isocontour within the function domain. Addressing this problem, by localizing the PCA strategy, is among our future goals. Moreover, in case of weak global structure benchmarks, PCA-BO might also benefit from a different scheme in the pre-tuning of the PCA algorithm: instead of weighting the sample set, a truncation scheme which only selects a prescribed percentage of samples -- ordered according to their function values --  to train the linear transformation parameters can be adopted.  
Within this study, we have assessed the performance of PCA-BO not only in terms of the empirical convergence rate, but also tracking the CPU time elapsed to complete the runs. Results showed that, except for the 10-variables test cases that can be efficiently addressed with standard BO, the average CPU time throughout the different tested functions can be consistently reduced by approaching the optimization problem with PCA-BO. In fact, training the GP model and optimizing the acquisition function is a much cheaper procedure in a search space of reduced dimensionality than in the full-dimensional space. 

In the future, we will extend our study by applying a non-linear transformation when tuning the parameters for the PCA-mapping to the lower-dimensional feature space. One option is represented by the Kernel Principal Component Analysis (KPCA)~\cite{scholkopf_nonlinear_1998}, which is a non-linear dimensionality reduction through the use of kernels, more capable to capture the isocontour of the cost functions when variables are not linearly correlated. Moreover, we will extend PCA-BO to make it capable of addressing real-world constrained applications, such as the optimization of mechanical structures subjected to static/dynamic loads. These problems are usually characterized by a great number of design variables and by costly evaluations, requiring optimization techniques of affordable computational cost.

\vspace{1ex} %important for final version only!
{\footnotesize{\textbf{Acknowledgments. } 
Our work was supported by the Paris Ile-de-France Region 
and by COST Action CA15140 ``Improving Applicability of Nature-Inspired Optimisation by Joining Theory and Practice (ImAppNIO)''.}}

\bibliographystyle{splncs04}
\bibliography{mybibliography}
\end{document}